%% file: main.tex
\documentclass[sigconf,natbib=true]{acmart}

\usepackage{enumitem}
\usepackage{hyperref}
\usepackage{multirow}
\usepackage{amsthm}
\usepackage{array}
\usepackage{subfigure}
\usepackage{graphicx}
\setlist[enumerate]{align=left}
\usepackage{algorithm}
\usepackage{algorithmic}

\usepackage[multiple]{footmisc}

\AtBeginDocument{%
  }

\setcopyright{acmlicensed}
\copyrightyear{2018}
\acmYear{2018}
\acmDOI{XXXXXXX.XXXXXXX}

\acmConference[Conference acronym 'XX]{Make sure to enter the correct
  conference title from your rights confirmation emai}{June 03--05,
  2018}{Woodstock, NY}

\acmISBN{978-1-4503-XXXX-X/18/06}

\begin{document}

\title{Benchmarking for Deep Uplift Modeling in Online Marketing}

\author{Dugang Liu}
\affiliation{%
  \institution{Guangdong Laboratory of Artificial Intelligence and Digital Economy (SZ)}
  \city{Shenzhen}
  \country{China}
}
\email{dugang.ldg@gmail.com}

\author{Xing Tang}
\author{Yang Qiao}
\affiliation{
  \institution{Financial Technology (FiT), Tencent}
  \city{Shenzhen}
  \country{China}
}
\email{xing.tang@hotmail.com}
\email{sunnyqiao@tencent.com}

\author{Miao Liu}
\affiliation{
  \institution{Shenzhen University}
  \city{Shenzhen}
  \country{China}
}
\email{miaoliu.lmm@gmail.com}

\author{Zexu Sun}
\affiliation{
  \institution{Renmin University of China}
  \city{Beijing}
  \country{China}
}
\email{sunzexu21@ruc.edu.cn}

\author{Xiuqiang He}
\affiliation{
  \institution{Financial Technology (FiT), Tencent}
  \city{Shenzhen}
  \country{China}
}
\email{xiuqianghe@tencent.com}

\author{Zhong Ming}
\affiliation{%
  \institution{Shenzhen Technology University}
  \city{Shenzhen}
  \country{China}
}
\email{mingz@szu.edu.cn}

\renewcommand{\shortauthors}{Dugang Liu, et al.}

\input{section/abstract}



\maketitle

\input{section/introduction}
\input{section/deep_uplift_modeling}

\input{section/benchmark}
\input{section/experiments}
\input{section/limitation}
\input{section/conclusion}

\bibliographystyle{ACM-Reference-Format}
\bibliography{sample-base}

\end{document}

%% file: section/abstract.tex
\begin{abstract}
Online marketing is critical for many industrial platforms and business applications, aiming to increase user engagement and platform revenue by identifying corresponding delivery-sensitive groups for specific incentives, such as coupons and bonuses. As the scale and complexity of features in industrial scenarios increase, deep uplift modeling (DUM) as a promising technique has attracted increased research from academia and industry, resulting in various predictive models. However, current DUM still lacks some standardized benchmarks and unified evaluation protocols, which limit the reproducibility of experimental results in existing studies and the practical value and potential impact in this direction. In this paper, we provide an open benchmark for DUM and present comparison results of existing models in a reproducible and uniform manner. To this end, we conduct extensive experiments on two representative industrial datasets with different preprocessing settings to re-evaluate 13 existing models. Surprisingly, our experimental results show that the most recent work differs less than expected from traditional work in many cases. In addition, our experiments also reveal the limitations of DUM in generalization, especially for different preprocessing and test distributions. Our benchmarking work allows researchers to evaluate the performance of new models quickly but also reasonably demonstrates fair comparison results with existing models. It also gives practitioners valuable insights into often overlooked considerations when deploying DUM. We will make this benchmarking library, evaluation protocol, and experimental setup available on GitHub.
\end{abstract}

%% file: section/introduction.tex
\section{Introduction}\label{sec:intro}
To increase user engagement and platform revenue, online marketing is an essential task for many industrial platforms and business applications, aiming to provide certain specific incentives to the target users, such as coupons~\cite{zhao2019unified}, discounts~\cite{lin2017monetary}, and bonuses~\cite{ai2022lbcf}.
The success of online marketing depends largely on being able to accurately identify the sensitive user groups for each incentive~\cite{he2022causal,xu2022learning}.
Uplift modeling is an effective tool to address this problem~\cite{devriendt2020learning,betlei2021uplift,chen2022imbalance}, which aims to estimate changes in outcomes due to individual-level treatments, i.e., accurately captures the difference between a user's response to various incentives and their responses without incentives.
This difference is called \textit{uplift}, and practitioners can naturally specify more effective marketing delivery strategies based on the estimated uplift.
Because uplift modeling typically targets a large user base, even a slight improvement in its predictive performance can significantly increase overall revenue.
For example, in Tencent's Licaitong financial scenario~\cite{liu2023explicit}, a slight absolute improvement in prediction performance will significantly reduce marketing costs and increase commission conversion.
Compared with response tasks like click-through rate prediction, uplift modeling tasks usually contain multiple incentives and lack direct supervision objectives.
In other words, we can only observe the user's response but not the degree of its change. 
Therefore, it is a great challenge to improve uplift modeling significantly.

Given the importance and unique challenges of uplift modeling, researchers in academia and industry have done extensive research in recent years.
Uplift prediction models have evolved from meta-learners-based~\cite{kunzel2019metalearners,nie2021quasi,bang2005doubly}, tree and forest-based~\cite{radcliffe2011real,chipman2010bart,wager2018estimation,athey2016recursive,wan2022gcf,ai2022lbcf}, Knapsack problem-based~\cite{goldenberg2020free,albert2021commerce} to deep neural networks-based architecture. 
Notably, many recent works have focused on developing new neural network architectures to better adapt to uplift modeling in industrial scenarios and shown remarkable performance improvements, such as EUEN~\cite{ke2021addressing},  DESCN~\cite{zhong2022descn} and EFIN~\cite{liu2023explicit} and so on.
We call this deep neural network-based line \textit{deep uplift modeling (DUM)}.
The trend of technical lines gradually moving closer to DUM largely follows the development of online marketing forms and models in industrial scenarios.
On the one hand, in real online marketing scenarios, incentives are becoming increasingly diverse and bring more helpful feature information (such as text and pictures displayed with bonuses, etc.). 
This implies the need to develop new techniques, including modeling the incentive-related features and the interaction between these features and the user and contextual features. 
DUM can more flexibly adapt to this goal than other lines.
On the other hand, since deep network-based models are more commonly deployed in industrial scenarios, it is also easier to integrate them with DUM techniques than other lines.
Therefore, we follow this trend and prioritize DUM in this benchmark work, which is more urgently needed in current industrial scenarios.

Despite the promising results from these aforementioned studies, there is a lack of standardized benchmarks and unified evaluation protocols for DUM tasks.
For example, although industrial benchmark datasets such as Criteo~\cite{diemert2021large} and Lazada~\cite{zhong2022descn} are used in existing studies, these works often use different random seeds to perform unknown train-test splits.
In addition, there is no consistent standard for data preprocessing steps, such as feature normalization, instance deduplication, etc, and discussion of their potential impacts is neglected.
This issue leads to non-reproducible and inconsistent experiment results between different studies and confuses practitioners.
Each study shows that they achieve better performance than state-of-the-art methods. 
However, results from the same model in different studies cannot be compared due to different evaluation protocols.
Due to the lack of open benchmarking results, practitioners may scrutinize whether the baseline model has been correctly implemented and rigorously tuned.
Only a few works present detailed implementations of baseline models, and some do not include open-source code. 
Therefore, in the research field of uplift modeling, especially for DUM tasks, practitioners are constantly forced to re-implement many baselines and evaluate the results on their own data partition without knowing their correctness, which leads to inefficient development and redundant work.
In addition, some factors that may need to be considered when deploying DUM have not received enough attention, such as the data preprocessing settings.

Inspired by the success of CausalML~\cite{chen2020causalml}, the most popular uplift modeling and causal inference benchmark and mainly focuses on traditional tree-based uplift modeling methods and rarely covers DUM, this paper proposes an open benchmarking framework for DUM, i.e., DUMOM.
Our work standardizes the open benchmarking pipeline for such modeling and thoroughly compares different DUM models for reproducible studies. 
We extensively evaluate 13 existing representative models in multiple data processing settings on two widely-used representative industrial datasets, including Criteo~\cite{diemert2021large} and Lazada~\cite{zhong2022descn}. 
The former contains training and test subsets with instance selection bias, and the latter contains a training subset with instance selection bias and a test subset without bias.
Therefore, we prioritize these two datasets in this benchmark work so that we can provide more comprehensive insights.
The comparison results between different models in our experiments reveal some valuable conclusions and can motivate attention to related research. 
Specifically, there is no stable state-of-the-art DUM model under different datasets and data preprocessing steps after sufficient hyperparameter search and model tuning.
In particular, when the dataset's training and test distributions are inconsistent, the differences between recent DUM models and traditional models are more minor than expected, and different DUM models are also sensitive to different data preprocessing steps.
In other words, current DUM research has apparent limitations in model generalization, and it is necessary to carefully consider the test distribution environment and appropriate data preprocessing when deploying DUM and conduct more research in these areas.

%% file: section/deep_uplift_modeling.tex
\section{Deep Uplift Modeling}\label{sec:deep_uplift_modeling}
In this section, we first define the uplift modeling task and architectural modules with the necessary notations and then briefly introduce some representative DUM models used in our experiments.
Finally, to avoid confusion, we emphasize the similarities and differences between the current DUM and related individual treatment effect (ITE) estimation studies.

\subsection{Architecture Overview}

\subsubsection{Objective of Uplift Modeling}
Uplift modeling seeks to determine the incremental impact of an intervention (or treatment) on an individual's behavior, where the intervention (or treatment) corresponds to an incentive in online marketing.
This is achieved by estimating an individual's potential outcome under different scenarios, such as with or without intervention. 
In this paper, without loss of generality, we use the binary treatment indicator $T\in \{0,1\}$, where $T=0$ represents the control group (i.e., without intervention) and $T=1$ represents the treatment group (i.e., with intervention). 
In line with the potential outcome framework~\cite{rubin2005causal}, each individual $i$ has two possible outcomes: $Y_i(0)$ in the absence of intervention and $Y_i(1)$ in the presence of intervention. 
The uplift of intervention $T$ on an individual $i$, represented by $\tau_i$, is equal to the difference between the two potential outcomes:
\begin{equation}\label{eq:1}
\tau_i =Y_i(1)-Y_i(0).
\end{equation}
However, individuals can only observe one of the two potential outcomes in real-world scenarios. 
The unobserved outcome is known as the counterfactual outcome, while the observed outcome can be represented as:
\begin{equation}\label{eq:2}
Y_i=T Y_i(1)+(1-T) Y_i(0).
\end{equation}
Since we can not observe both potential outcomes for each individual $i$, the uplift cannot be directly estimated.
However, with some adequate assumptions, it can be obtained in a manner similar to the conditional average treatment effect (CATE), which is the average treatment effect conditioning on a set of covariates. 
Specially, let $X=\boldsymbol{x}$ be a $k$-dimensional covariate vector, then the CATE $\tau(\boldsymbol{x})$ can be derived as:
\begin{equation}\label{eq:3}
\tau(\boldsymbol{x})=\mathbb{E}[\tau \mid X=\boldsymbol{x}]=\mathbb{E}[Y(1)-Y(0) \mid X=\boldsymbol{x}].
\end{equation}
The objective of uplift modeling in Eq.\eqref{eq:3} involves estimating two conditional expectations of the observed outcomes, which is feasible with available data.
Next, we describe the critical components of a DUM model architecture.

\subsubsection{Response Modeling of the Control Group}
Based on the input covariate vector $\mathbf{x}$, DUM usually transforms it into a low-dimensional dense vector $\mathbf{e}$ through an embedding layer.
Then, a network architecture with predictive capabilities will be used to obtain the response of the control group $\hat{Y}_i(0)$ from this low-dimensional dense vector.

\subsubsection{Response Modeling of the Treatment Group}
Similarly, there is usually another network architecture with predictive capabilities responsible for predicting the response of the treatment group $\hat{Y}_i(1)$.
Note that in some DUM models, a shared module shares information at the bottom of the above two prediction networks.

\subsubsection{Predictive Modeling of Treatment Indicator Variables}
Since accurately predicting whether a user belongs to the control group or the treatment group helps to make the model robust to the distribution imbalance of the group to a certain extent, a crucial technical branch in DUM is whether to additionally predict the indicator variable of the treatment $\hat{T}_i$.
An independent prediction network will perform this process.

\subsubsection{Loss Function}
The least-square error (MSE) loss is widely used in DUM tasks, which is defined as follows:
\begin{equation}\label{equ:4}
\mathcal{L} = \frac{1}{N}\sum_{\mathcal{D}}\left(Y_i- \hat{Y}_i\right)^2,
\end{equation}
where $\mathcal{D}$ is the training set with $N$ instances. 
Depending on the group to which each instance $X$ belongs, the prediction of $Y_i$ may be generated by $\hat{Y}_i(0)=f(\mathbf{e}|T_i=0,\theta_0)$ or $\hat{Y}_i(1)=f(\mathbf{e}|T_i=1,\theta_1)$, where $\theta_0$ and $\theta_1$ are respective model parameters. 
Note that for brevity, we omit the optimization loss term for $T$.

\subsubsection{Inference Stage}
In practical applications, DUM pays more attention to the response changes that each incentive may bring to the user than the response itself.
Therefore, in the inference stage, we will make predictions for each instance based on the trained model on its response in the control group and the intervention group and then use the difference between the two to perform a ranking of delivery strategies, i.e.,
\begin{equation}\label{equ:5}
\hat{\tau_i} = \hat{Y}_i(1)-\hat{Y}_i(0).
\end{equation}

\subsection{Representative Models}
\begin{figure*}[htbp]
    \centering
    \includegraphics[width=0.9\textwidth]{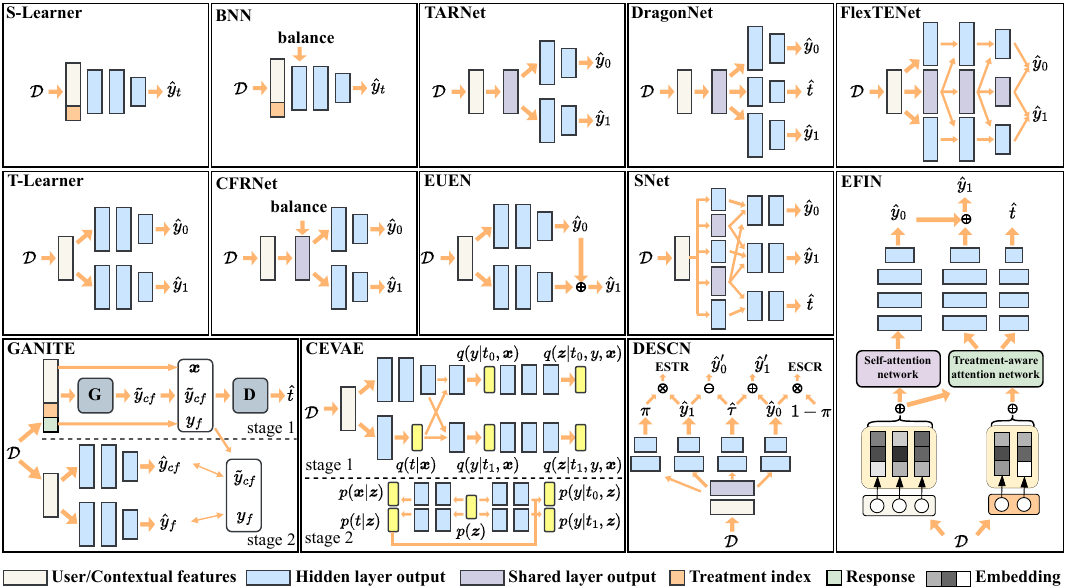}
    \caption{The architectural illustration of representative baselines in neural network-based uplift modeling. For more information on the meaning of some baseline-specific symbols, please refer to their original papers.}
    \label{fig:baseline}
\end{figure*}

Next, we summarize the representative models evaluated and benchmarked in this work.
Note that although we only enumerate a subset of existing DUM models, they cover a broad range of representative research lines on DUM. 
To facilitate understanding and comparison, we show the architecture diagrams of these representative models in Figure~\ref{fig:baseline}.
As shown in Figure~\ref{fig:baseline}, at a high level, we divide the existing methods into two categories based on how the treatment indicator is used.

\subsubsection{Treatment as a Branch Switch} 
Since it is necessary to determine the group to which each instance belongs during loss optimization, the corresponding prediction branch is selected to obtain the prediction label.
Therefore, the methods of this category only use the treatment indicator as a switch to switch the model branch and do not let it participate in the model's training.
Representative models in this category include T-Learner~\cite{kunzel2019metalearners}, TARNet~\cite{shalit2017estimating}, CFRNet~\cite{shalit2017estimating}, FlexTENet~\cite{curth2021inductive}, and EUEN~\cite{ke2021addressing}.

\subsubsection{Treatments as Model Features}
Since the intervention itself may also carry helpful information or needs to be exploited to perform some steps that help alleviate the distribution shift, the methods in this category incorporate the treatment indicators as features into the model's training.
Representative models in this category include BNN~\cite{johansson2016learning}, S-Learner~\cite{kunzel2019metalearners}, DragonNet~\cite{shi2019adapting}, SNet~\cite{curth2021nonparametric}, GANITE~\cite{yoon2018ganite}, CEVAE~\cite{louizos2017causal}, DESCN~\cite{zhong2022descn}, and EFIN~\cite{liu2023explicit}

\subsubsection{Overview of Representative Models}
For models in the first category,
\begin{itemize}[leftmargin=*]
    \item \textbf{T-Learner}~\cite{kunzel2019metalearners}: it is similar to S-Learner, which uses two estimators for the treatment and control groups.
    \item \textbf{TARNet}~\cite{shalit2017estimating}: a deep learning-based method that extends the two-model approach, which can be seen as the extension of BNN.   
    \item \textbf{CFRNet}~\cite{shalit2017estimating}: it is built upon the TARNet, which adds the minimization of the discrepancy measured by either the maximum mean discrepancy or the Wasserstein distance between the two distributions.
    \item \textbf{FlexTENet}~\cite{curth2021inductive}: it builds on ideas from multi-task learning to design a new architecture for CATE estimation, which adaptively learns what to share between the potential outcome functions.
    \item \textbf{EUEN}~\cite{ke2021addressing}: it is an uplift model for large-scale online advertising, which uses two sub-networks to estimate the control conversion probability and the uplift effect, respectively.
\end{itemize}
For models in the second category,
\begin{itemize}[leftmargin=*]
    \item \textbf{BNN}~\cite{johansson2016learning}: it leverages deep learning for counterfactual inference with theoretical guarantees, combining ideas from domain adaptation and representation learning.
    \item \textbf{S-Learner}~\cite{kunzel2019metalearners}: it is a kind of meta-learner method that uses a single estimator to estimate the outcome without giving the treatment a special role. Varying the treatment while fixing other features yields different variants and uses their differences to estimate CATE.  
    \item  \textbf{DragonNet}~\cite{shi2019adapting}: it designs a new architecture with the propensity score for estimation adjustment and uses a targeted regularization procedure to induce a bias toward models.
    \item \textbf{SNet}~\cite{curth2021nonparametric}: it is a new classification of meta-learners inspired by the ATE estimator taxonomy, which estimates unbiased pseudo-outcomes based on regression adjustment, propensity weighting, or both.
    \item \textbf{GANITE}~\cite{yoon2018ganite}: it utilizes a generative adversarial network (GAN) to model treatment effect heterogeneity, which consists of two components, a counterfactual block that generates the counterfactual outcome and an ITE block that generates the ITE.
    \item \textbf{CEVAE}~\cite{louizos2017causal}: it is a variational autoencoder (VAE)-based treatment effect heterogeneity modeling method, which uses a VAE to learn a latent confounding set from the observed covariates and then uses it to estimate the treatment effect.
    \item \textbf{DESCN}~\cite{zhong2022descn}: it jointly learns the treatment and response functions in the entire sample space to avoid treatment bias and employs an intermediate pseudo treatment effect prediction network to relieve sample imbalance.
    \item \textbf{EFIN}~\cite{liu2023explicit}: it is a feature interaction-aware uplift method, which consists of four customized modules: a feature encoding module, a self-interaction module, a treatment-aware interaction module, and an intervention-constrained module.
\end{itemize}

\subsection{The Connection between DUM and ITE}
Because both aimed to understand how treatments affect individuals, early developments in DUM borrowed heavily from ideas about estimating individual treatment effects (ITE) in causal inference, although they focus on marketing campaigns and medical treatments, respectively.
However, the development of the online marketing scenario has made the research and goals in DUM more unique, and we argue that they already differ significantly in evaluation metrics, feature complexity, and treatment types.

1) \textbf{The Evaluation Metrics.}
DUM focuses on accurately distinguishing the response difference estimates of user groups that tend to respond from other groups, i.e., ranking their response differences as high as possible, without paying attention to the specific values of the corresponding differences.
ITE focuses on estimating the specific difference in potential outcomes for each individual with and without treatment.
Therefore, DUM tends to discover the user groups that tend to convert, and ITE aims to obtain accurate ITE results on the individual level.
The core metrics are the Area Under Uplift Curve (AUUC) and Qini coefficient (QINI)~\cite{betlei2021uplift}, etc., which are different from the commonly used $\epsilon_{PEHE}$ and $\epsilon_{ATE}$~\cite{shalit2017estimating} in ITE.
These metrics draw on ranking metrics from traditional click-through rate prediction tasks, such as AUC~\cite{zhu2021open}.
2) \textbf{The Features Complexity.}
DUM is particularly suitable for optimizing resource allocation, for instance, determining which customers are most likely to increase purchases due to a marketing campaign.
This requirement may invoke the deployment of learned models capable of navigating the high-dimensional feature space to capture the diverse impacts of features on the uplift prediction.
ITE is more focused on personalized decision-making, such as medical plans.
Therefore, it typically does not consider too many feature dimensions, and the model's training process is more focused on finding which features best identify individuals who will benefit.
3) \textbf{The Treatment Type.}
DUM's flexibility lies in its ability to handle various treatment types.
This means it can assess not only single binary treatments (e.g., whether a customer received a marketing email) but also analyze the impact of different treatments (e.g., various marketing strategies, product discounts) on individual behavior. 
In addition, DUM can be better compatible with displaying discount coupons in marketing scenarios where treatments often contain helpful information, such as text and images.
In contrast, ITE estimates usually focus on simple treatment situations, such as binary treatments (e.g., whether a patient received a drug).
Providing deep insights into different types and degrees of treatment effects like DUM is not easy.
Considering these unique features of DUM, the DUM benchmark will differ from some existing ITE benchmarks in different dimensions and should not be conflated, especially the adopted datasets, baseline methods, and evaluation metrics.

%% file: section/benchmark.tex
\section{Open DUM Benchmark (DUMOM)}\label{sec:benchmark}
Although the various DUM models mentioned above have achieved promising results, current research lacks standardized benchmarks and unified evaluation protocols.
These problems cause most studies to face obstacles of non-reproducibility and inconsistent results, hindering the progress of DUM model development.
In this section, We first compare our work to other related benchmarks or libraries to illustrate the necessity of an open benchmark for DUM. 
Then, we provide the implementation details of our open DUM benchmark.

\subsection{Comparison with Existing Work}
Based on a review of existing research, we identify several essential requirements that impact the reproducibility and value of DUM benchmarks.
These include three aspects that affect the difficulty of use: benchmark results, hyperparameter search, and optimal parameters provided; two aspects of data preprocessing settings, instance deduplication and feature normalization; and one aspect that measures the degree of compatibility with DUM.
As shown in Table~\ref{tab:comparison}, we evaluate a set of existing benchmarks and libraries for uplift modeling and causal inference that are powerfully relevant to our work against these considerations, including Criteo-ITE-Benchmark~\cite{diemert2021large}, CATENets~\cite{curth2021nonparametric},  DoWhy~\cite{dowhy}, EconML~\cite{econml}, CausalML~\cite{chen2020causalml}, DECI\footnote{ https://github.com/microsoft/causica}, ShowWhy\footnote{ https://github.com/microsoft/showwhy}. 
We find that none of them fully meet these requirements, which motivates us to propose a new public benchmark for DUM in this paper.

\begin{table*}[htbp]
\caption{Comparison with other relevant benchmarks or libraries on reproducible configurations, where $\checkmark$\,|\,\textendash\,|\,$\times$ mean fully supported, partially supported, and not supported, respectively.}
\centering
\scalebox{1.0}{
\begin{tabular}{l|cccccccc}
\specialrule{0.1em}{1pt}{1pt}
        Benchmark or Libararys & Criteo & CATENets & DoWhy & EconML & causalML & DECI & ShowWhy & Ours \\ 
        \hline
        Benchmark Results & $\checkmark$ & \textendash & \textendash & \textendash & \textendash & $\times$ & $\times$ & $\checkmark$\\
        Hyperparameter Search & $\checkmark$ & $\times$ & $\checkmark$ & $\checkmark$ & $\checkmark$ & $\times$ & $\times$ & $\checkmark$\\
        Optimal Parameters & $\times$ & $\times$ & $\times$ & $\times$ & $\times$ & $\times$ & $\times$ & $\checkmark$\\
        Feature Normalization & $\checkmark$ & $\checkmark$ & $\checkmark$ & $\checkmark$ & $\checkmark$ & $\times$ & $\times$ & $\checkmark$\\
        Instance Deduplication & $\times$ & $\times$ & $\times$ & $\times$ & $\times$ & $\times$ & $\times$ & $\checkmark$\\
        Comprehensive DUM Compatibility & \textendash & \textendash & \textendash & \textendash & \textendash & $\times$ & $\times$ & $\checkmark$\\
\specialrule{0.1em}{1pt}{1pt}
\end{tabular}}
\label{tab:comparison}
\end{table*}

Specifically, for the first three aspects: 
1) \textit{Benchmark results.} Although many uplift modeling studies provide partial baseline results, they often lack important details such as baseline model implementation, training code, and hyperparameters. Furthermore, they have insufficient models, datasets, and evaluation metrics coverage. 
2) \textit{Hyperparameter search.} A thorough hyperparameter search is a prerequisite for evaluating the model's true performance. Most existing benchmarks or libraries provide hyperparameter search capabilities based on \textit{scikit-learn} or other toolkits. However, they do not provide a reasonable hyperparameter search range, which makes it difficult for researchers to find the optimal parameters for different models quickly.
3) \textit{Optimal parameters provided.} Providing optimal parameters can speed up model replication for practitioners and prevent wasted resources and inconsistent results. Unfortunately, most existing libraries do not provide optimal parameters, which complicates reproducibility. 

For the latter three aspects: 
1) \textit{Instance deduplication.} In particular, we note that existing industrial datasets often contain duplicate instances, i.e., users with the same profile appear multiple times in the dataset. Our empirical validation shows that removing these duplicates can significantly impact performance evaluation. Unfortunately, this is often overlooked by existing research and libraries. 
2) \textit{Feature normalization.} Feature processing plays a crucial role in the performance of the model. Our analysis shows that feature normalization has a significant impact on improving estimation results. However, many existing libraries provide some feature transformation function modules but ignore key details, such as how to handle numerical features and whether to perform feature normalization. 
3) \textit{Comprehensive DUM compatibility.} Different from other routes, the DUM model requires various complex deep network modules, and there are many reusable structures. Therefore, the targeted development of a set of pluggable and reusable network components similar to those processed in traditional click-through rate prediction benchmarks~\cite{zhu2021open} is necessary and beneficial for DUM. However, since existing libraries or benchmarks do not pay much attention to DUM, they lack sufficient support.

\subsection{Evaluation Protocol}
Next, we provide a detailed introduction to the evaluation protocol adopted in our benchmark.

\subsubsection{Datasets} 
Our experiments adopt two representative industrial datasets to better adapt to different marketing scenarios: 1) Lazada~\cite{zhong2022descn}. It is a dataset collected on the Lazada platform, containing 83 continuous features, a processing indicator, and a label. It provides data sets from two different sources, where the training set is collected from a production environment with a treatment bias, the treatment allocation is selective due to the operation target policy, and the test set is collected from the users who are not affected by the targeting strategy and the treatment assignment follows the randomized controlled trials (RCT). 2) Criteo~\cite{diemert2021large} is a dataset from real advertising scenarios provided by Criteo AI Labs. It contains nearly 14 million instances with similar treatment bias, 12 continuous features, a treatment indicator, and 2 labels (i.e., visits and conversions). 
The statistics of the two public datasets are shown in Table~\ref{tab:datasets}.
\begin{table}[htbp]
\caption{Statistics of the two public datasets.}
\centering
\scalebox{0.85}{
\begin{tabular}{l|l|l|l}
\specialrule{0.1em}{1pt}{1pt}
\textbf{Dataset} & \textbf{Criteo} & \multicolumn{2}{c}{\textbf{Lazada}} \\
\hline
Type & Train & Train & Test \\
\hline
Size & 13,979,592 & 926,669 & 181,669 \\
Ratio of Treatment to Control & 5.67:1 & 0.28:1 & 1.09:1\\
Positive Sample Ratio & 4.70\% & 1.99\% & 3.52\%\\
Relative Average Uplift & 27.07\% & 499.8\% & 11.2\%\\
Average Uplift & 1.03\% & 4.72\% & 0.37\%\\
Features Number & 12 & 83 & 83\\
Conversion Target & Visit & Label & Label\\

\specialrule{0.1em}{1pt}{1pt}
\end{tabular}}
\label{tab:datasets}
\end{table}

\subsubsection{Data Splitting}
Following most existing studies~\cite{liu2023explicit}, we randomly split the Criteo dataset into three sets, with a ratio of 8:1:1 for the training, validation, and testing.
For the Lazada dataset, to make it accurately reproducible and easy to compare with existing work, we follow the splitting strategy reported by DESCN~\cite{zhong2022descn}: we randomly split the training set into two subsets with a ratio of 9:1 for training and validation, and the RCT test data is directly used as the test set. 
We set the random seed to 0-9, respectively, to ensure reproducibility when performing random splits.

\subsubsection{Data Preprocessing}
Since all feature values are continuous and do not require any special processing, we use all features from both datasets.
Considering that our benchmark investigates the effect of data normalization on the results, we perform batch normalization~\cite{ioffe2015batch} to the input features when choosing to perform normalization on the dataset.
The expression can be referenced with Eq.\eqref{equ:6}.
\begin{equation}\label{equ:6}
    \hat{x}_i = \frac{x_i - E[x]}{\sqrt{Var[x] + \epsilon}}\rightarrow y_i = \gamma \hat{x}_i + \beta,
\end{equation}
where $\gamma$ is the scaling factor and $\beta$ is the bias term. 
During the network's training process, both $\gamma$ and $\beta$ are learnable parameters that can be updated by the back-propagation algorithm.

\subsubsection{Evaluation Metrics}
To evaluate each model's effectiveness comprehensively, we adopt four widely used metrics in DUM: AUUC, QINI, WAU, and the uplift score at the first $k$ percentile (LIFT@$k$).
We report the results with $k$ set to 30 and compute these metrics using a standard python package \textit{scikit-uplift}\footnote{https://www.uplift-modeling.com}.   
  
\subsubsection{Benchmark Models and Implementation Details} 
We select all representative methods described in Section~\ref{sec:deep_uplift_modeling} as benchmark models.
Because PyTorch can avoid non-determinism on GPU devices better than Tensorflow~\cite{zhu2021open}, we choose it to implement all models.
We utilized the AdamW optimizer and set a maximum of 20 iterations.
Our model training is performed on V100 GPU clusters.
Due to the different workloads of the GPU cluster, each model's epoch training time is for reference only. 
We record all data settings and model hyperparameters in configuration files to make configurations more convenient. 
To promote more reproducible research, we plan to open-source the implementation and settings of all models to the community after the paper is accepted.
    
\subsubsection{Hyperparameter Tuning}
To ensure a fair and objective evaluation of all models' true performance, we conduct a thorough search for hyperparameters using QINI as the main evaluation metric. 
In addition, we adopt an early-stopping mechanism with a patience of 5 to avoid overfitting the training set. 
To improve efficiency, we use the hyperparameter search library \textit{Optuna}\footnote{https://optuna.org/} to speed up the tuning process. 
The search range of all hyperparameters is shown in Table~\ref{tab:search_ranges}.
Note that different models may have different numbers of auxiliary losses, and we use the same parameter search range across all these losses.

\begin{table}[htbp]
\caption{Hyperparameters and their values tuned in the experiments.}
\centering
\scalebox{0.85}{
\begin{tabular}{ccc}
\specialrule{0.1em}{3pt}{3pt}
\textbf{Name} & \textbf{Range} & \textbf{Functionality}\\
\specialrule{0.05em}{3pt}{3pt}
$rank$ & $\left\{2^{5},2^{6},2^{7}\right\}$ & Hidden units \\
$bs$ &$\left\{2^{8},2^{9},2^{10},2^{11}\right\}$ & Batch size \\
$lr$ &$\left\{1e^{-4},5e^{-4},1e^{-3},5e^{-3},1e^{-2}\right\}$ & Learning rate \\
$\lambda$ &$\left\{1e^{-5},1e^{-4},1e^{-3},1e^{-2}\right\}$ & Weight decay 
\\
$\alpha$ &$\left\{0.1, 0.2, 0.3, 0.4, 0.5, 0.6, 0.7, 0.8, 0.9\right\}$ & Auxiliary losses weight
\\
\specialrule{0.1em}{3pt}{3pt}\\
\end{tabular}}
\vspace{-15pt}
\label{tab:search_ranges}
\end{table}

%% file: section/experiments.tex
\section{Experiment Results and Discussion}\label{sec:experiment}
In this section, we provide valuable discussions and insights into the benchmark results and offer new and instructive experiences for the practical deployment of DUM.
Again, we emphasize that the training and testing distributions are not significantly inconsistent on Criteo, but the opposite is true on Ladaza.

\subsection{Full Benchmark Results}\label{appendix:result}
Benchmark results using different data processing combinations on all datasets are shown in Tables~\ref{tab:result_1},~\ref{tab:result_2},~\ref{tab:result_3}, and~\ref{tab:result_4}.
\begin{table*}[htbp]
\caption{Benchmark results on Criteo and Lazada dataset with instance deduplication and without feature normalization. We highlight the top 3 best results, where the best results are marked in bold.}
    \centering
    \resizebox{0.89\textwidth}{!}{
    \begin{tabular}{c|c|c|cccc|cccc|ccc}
    \specialrule{0.1em}{3pt}{3pt}
    
     \multirow{2}*{Dataset} &
     \multirow{2}*{Year} & 
     \multirow{2}*{Model} & 
     \multicolumn{4}{c|}{Valid Metrics} & 
     \multicolumn{4}{c|}{Test Metrics} & 
     \multicolumn{3}{c}{Train Infos} \\
     \cline{4-14} 
       & & &QINI & AUUC & WAU & LIFT@30 & QINI & AUUC & WAU & LIFT@30 & Time(s) & Epochs & Params \\ 
     \hline
        
    \multirow{13}{*}{Criteo}  
        & 2012 & S-Learner & 0.0946 & 0.0370 & 0.0089 & 0.0329 & 0.0951 & 0.0375 & 0.0087 & 0.0326 & 34 & 4 & 28697 \\ 
        & 2016 & T-Learner & 0.1035 & 0.0407 & 0.0093 & 0.0334 & 0.1053 & 0.0416 & 0.0091 & 0.0334 & 71 & 1 & 4314 \\ 
        & 2016 & BNN & 0.1047 & 0.0411 & \underline{0.0098} & 0.0341 & 0.1017 & 0.0402 & \underline{0.0098} & 0.0333 & 36 & 13 & 2201 \\ 
        & 2017 & TARNet & 0.1024 & 0.0406 & 0.0090 & 0.0439 & 0.1063 & 0.0425 & 0.0087 & 0.0442 & 111 & 10 & 14426 \\ 
        & 2017 & CFRNet & 0.1075 & 0.0426 & 0.0087 & 0.0472 & 0.1085 & 0.0434 & 0.0088 & 0.0466 & 212 & 7 & 14426 \\ 
        & 2017 & CEVAE & 0.0975 & 0.0383 & \textbf{0.0112} & 0.0355 & 0.0881 & 0.0348 & \textbf{0.0111} & 0.0341 & 113 & 12 & 99270 \\ 
        & 2018 & GANITE & \underline{0.1209} & \underline{0.0482} & 0.0097 & \underline{0.0489} & \underline{0.1192} & \underline{0.0478} & 0.0097 & \underline{0.0474} & 331 & 12 & 139933 \\ 
        & 2019 & DragonNet & 0.1085 & 0.0430 & 0.0087 & 0.0460 & 0.1099 & 0.0439 & 0.0089 & 0.0463 & 133 & 7 & 18652 \\ 
        & 2021 & FlexTENet & 0.1146 & \underline{0.0460} & 0.0093 & \textbf{0.0496} & \underline{0.1164} & \underline{0.0470} & 0.0094 & \textbf{0.0494} & 72 & 3 & 18619 \\ 
        & 2021 & SNet & 0.1016 & 0.0401 & 0.0086 & 0.0407 & 0.1044 & 0.0414 & 0.0083 & 0.0405 & 61 & 4 & 61339 \\ 
        & 2021 & EUEN & \underline{0.1158} & 0.0459 & 0.0086 & 0.0462 & 0.1143 & 0.0456 & 0.0087 & 0.0448 & 63 & 9 & 15292 \\ 
        & 2022 & DESCN & 0.0817 & 0.0316 & 0.0086 & 0.0335 & 0.0795 & 0.0309 & 0.0085 & 0.0321 & 72 & 10 & 25437 \\ 
        & 2023 & EFIN & \textbf{0.1255} & \textbf{0.0502} & \underline{0.0110} & \underline{0.0478} & \textbf{0.1220} & \textbf{0.0491} & \underline{0.0110} & \underline{0.0473} & 517 & 11 & 39950 \\ \hline
        
        \multirow{13}{*}{Lazada} 
        & 2012 & S-Learner & 0.4192 & 0.0013 & 0.0292 & 0.0470 & 0.0178 & 0.0024 & 0.0037 & \underline{0.0082} & 7 & 4 & 12391 \\ 
        & 2016 & T-Learner & 0.4431 & 0.0076 & 0.0372 & 0.0533 & \underline{0.0231} & \underline{0.0032} & \textbf{0.0041} & 0.0064 & 7 & 5 & 75432 \\ 
        & 2016 & BNN & 0.4741 & 0.0064 & 0.0394 & 0.0579 & 0.0199 & 0.0028 & 0.0037 & 0.0077 & 19 & 10 & 12391 \\ 
        & 2017 & TARNet & 0.4394 & 0.0047 & 0.0369 & 0.0541 & 0.0177 & 0.0024 & \underline{0.0040} & 0.0051 & 7 & 8 & 64680 \\ 
        & 2017 & CFRNet & 0.4595 & 0.0084 & \underline{0.0405} & 0.0585 & \underline{0.0234} & \underline{0.0032} & 0.0036 & \underline{0.0079} & 9 & 11 & 64680 \\ 
        & 2017 & CEVAE & 0.4424 & 0.0049 & 0.0386 & 0.0536 & 0.0078 & 0.0010 & 0.0035 & 0.0052 & 35 & 13 & 234226 \\ 
        & 2018 & GANITE & 0.4946 & \underline{0.0138} & 0.0194 & \underline{0.0680} & 0.0187 & 0.0025 & 0.0035 & 0.0075 & 28 & 12 & 50475 \\ 
        & 2019 & DragonNet & \underline{0.4961} & 0.0110 & 0.0386 & 0.0604 & 0.0177 & 0.0024 & 0.0038 & 0.0069 & 9 & 11 & 81322 \\ 
        & 2021 & FlexTENet & \underline{0.4995} & 0.0110 & \underline{0.0402} & \underline{0.0662} & 0.0215 & 0.0030 & 0.0036 & \textbf{0.0085} & 21 & 10 & 166377 \\ 
        & 2021 & SNet & 0.4502 & \textbf{0.0185} & 0.0310 & 0.0652 & 0.0190 & 0.0026 & 0.0039 & 0.0076 & 20 & 6 & 157353 \\ 
        & 2021 & EUEN & 0.4513 & 0.0075 & 0.0377 & 0.0550 & 0.0205 & 0.0028 & \underline{0.0040} & 0.0076 & 7 & 4 & 16770 \\ 
        & 2022 & DESCN & 0.4671 & 0.0070 & 0.0357 & 0.0533 & \textbf{0.0252} & \textbf{0.0034} & 0.0038 & 0.0077 & 8 & 3 & 30123 \\ 
        & 2023 & EFIN & \textbf{0.5181} & \underline{0.0117} & \textbf{0.0433} & \textbf{0.0771} & 0.0169 & 0.0023 & 0.0036 & 0.0074 & 152 & 11 & 186754 \\ 
        \specialrule{0.1em}{3pt}{3pt}
    \end{tabular}}
    \label{tab:result_1}
\end{table*}

\begin{table*}[htbp]
\caption{Benchmark results on Criteo and Lazada dataset with instance deduplication and feature normalization. We highlight the top 3 best results, where the best results are marked in bold.}
    \centering
    \resizebox{0.89\textwidth}{!}{
    \begin{tabular}{c|c|c|cccc|cccc|ccc}
    \specialrule{0.1em}{3pt}{3pt}
    
     \multirow{2}*{Dataset} &
     \multirow{2}*{Year} & 
     \multirow{2}*{Model} & 
     \multicolumn{4}{c|}{Valid Metrics} & 
     \multicolumn{4}{c|}{Test Metrics} & 
     \multicolumn{3}{c}{Train Infos} \\
     \cline{4-14} 
       & & &QINI & AUUC & WAU & LIFT@30 & QINI & AUUC & WAU & LIFT@30 & Time(s) & Epochs & Params \\ 
     \hline
        
    \multirow{13}{*}{Criteo}  
        & 2012 & S-Learner & 0.0885 & 0.0342 & 0.0085 & 0.0332 & 0.0825 & 0.0321 & 0.0083 & 0.0320 & 97 & 15 & 7705 \\ 
        & 2016 & T-Learner & 0.1029 & 0.0406 & 0.0083 & 0.0429 & 0.1034 & 0.0411 & 0.0082 & 0.0421 & 68 & 6 & 15258 \\ 
        & 2016 & BNN & 0.0939 & 0.0369 & 0.0097 & 0.0335 & 0.0911 & 0.0360 & 0.0096 & 0.0323 & 191 & 3 & 2201 \\ 
        & 2017 & TARNet & 0.1107 & 0.0438 & 0.0083 & 0.0442 & 0.1087 & 0.0433 & 0.0081 & 0.0434 & 207 & 6 & 55450 \\ 
        & 2017 & CFRNet & 0.1140 & 0.0454 & 0.0088 & 0.0485 & \underline{0.1194} & \underline{0.0478} & 0.0088 & \underline{0.0485} & 43 & 5 & 3898 \\ 
        & 2017 & CEVAE & 0.0315 & 0.0124 & \textbf{0.0148} & 0.0217 & 0.0334 & 0.0132 & \textbf{0.0146} & 0.0231 & 194 & 1 & 26238 \\ 
        & 2018 & GANITE & \textbf{0.1286} & \textbf{0.0514} & \underline{0.0101} & \textbf{0.0499} & \textbf{0.1272} & \textbf{0.0512} & \underline{0.0100} & \textbf{0.0491} & 558 & 7 & 139933 \\ 
        & 2019 & DragonNet & 0.1073 & 0.0423 & 0.0088 & 0.0379 & 0.1057 & 0.0420 & 0.0088 & 0.0371 & 98 & 2 & 72092 \\ 
        & 2021 & FlexTENet & 0.1110 & 0.0441 & 0.0087 & 0.0463 & 0.1125 & 0.0450 & 0.0084 & 0.0459 & 71 & 8 & 18619 \\ 
        & 2021 & SNet & 0.1118 & 0.0444 & 0.0089 & 0.0460 & 0.1127 & 0.0451 & 0.0088 & 0.0456 & 74 & 7 & 125403 \\ 
        & 2021 & EUEN & \underline{0.1195} & \underline{0.0475} & \underline{0.0101} & \underline{0.0492} & 0.1186 & 0.0475 & 0.0099 & \underline{0.0486} & 67 & 8 & 15292 \\ 
        & 2022 & DESCN & 0.0906 & 0.0360 & 0.0100 & 0.0382 & 0.0827 & 0.0330 & \underline{0.0100} & 0.0358 & 118 & 6 & 25437 \\ 
        & 2023 & EFIN & \underline{0.1238} & \underline{0.0491} & 0.0096 & \underline{0.0488} & \underline{0.1208} & \underline{0.0483} & 0.0095 & 0.0476 & 547 & 15 & 21216 \\ \hline
        
        \multirow{13}{*}{Lazada} 
        & 2012 & S-Learner & 0.4022 & 0.0006 & 0.0277 & 0.0451 & \underline{0.0246} & 0.0034 & 0.0035 & 0.0079 & 7 & 3 & 37927 \\ 
        & 2016 & T-Learner & 0.4348 & 0.0047 & 0.0359 & 0.0534 & 0.0163 & 0.0022 & \underline{0.0038} & 0.0066 & 13 & 1 & 75432 \\ 
        & 2016 & BNN & 0.4583 & 0.0045 & 0.0379 & 0.0526 & \underline{0.0292} & \underline{0.0040} & \textbf{0.0039} & 0.0078 & 8 & 3 & 37927 \\ 
        & 2017 & TARNet & 0.4527 & 0.0070 & 0.0370 & 0.0569 & 0.0266 & \underline{0.0037} & \underline{0.0038} & 0.0078 & 20 & 1 & 6312 \\ 
        & 2017 & CFRNet & 0.4714 & 0.0073 & 0.0357 & 0.0561 & \textbf{0.0303} & \textbf{0.0042} & 0.0037 & \textbf{0.0090} & 6 & 2 & 64680 \\ 
        & 2017 & CEVAE & 0.4388 & 0.0066 & \underline{0.0398} & \underline{0.0579} & 0.0121 & 0.0017 & 0.0037 & 0.0064 & 31 & 4 & 201330 \\ 
        & 2018 & GANITE & \underline{0.4944} & 0.0089 & 0.0241 & 0.0572 & 0.0194 & 0.0026 & 0.0036 & 0.0074 & 41 & 4 & 167339 \\ 
        & 2019 & DragonNet & \underline{0.5023} & \textbf{0.0173} & \underline{0.0432} & \textbf{0.0766} & 0.0209 & 0.0029 & \underline{0.0038} & 0.0079 & 14 & 3 & 23338 \\ 
        & 2021 & FlexTENet & 0.4413 & 0.0079 & 0.0326 & 0.0549 & 0.0220 & 0.0030 & 0.0036 & \underline{0.0088} & 9 & 10 & 60009 \\ 
        & 2021 & SNet & 0.4371 & 0.0035 & 0.0335 & 0.0523 & 0.0231 & 0.0032 & 0.0035 & \underline{0.0089} & 13 & 3 & 132905 \\ 
        & 2021 & EUEN & 0.4528 & 0.0075 & 0.0384 & 0.0580 & 0.0206 & 0.0028 & 0.0036 & 0.0076 & 7 & 1 & 50010 \\ 
        & 2022 & DESCN & 0.4734 & \underline{0.0104} & 0.0266 & 0.0573 & 0.0197 & 0.0027 & 0.0036 & 0.0069 & 9 & 6 & 9131 \\ 
        & 2023 & EFIN & \textbf{0.5510} & \underline{0.0148} & \textbf{0.0446} & \underline{0.0748} & 0.0131 & 0.0018 & 0.0037 & 0.0084 & 83 & 12 & 376914 \\ 
        
        \specialrule{0.1em}{3pt}{3pt}
    \end{tabular}}
    \label{tab:result_2}
\end{table*}

\begin{table*}[htbp]
\caption{Benchmark results on Criteo and Lazada dataset without instance deduplication and feature normalization. We highlight the top 3 best results, where the best results are marked in bold.}
    \centering
    \resizebox{0.89\textwidth}{!}{
    \begin{tabular}{c|c|c|cccc|cccc|ccc}
    \specialrule{0.1em}{3pt}{3pt}
    
     \multirow{2}*{Dataset} &
     \multirow{2}*{Year} & 
     \multirow{2}*{Model} & 
     \multicolumn{4}{c|}{Valid Metrics} & 
     \multicolumn{4}{c|}{Test Metrics} & 
     \multicolumn{3}{c}{Train Infos} \\
     \cline{4-14} 
       & & &QINI & AUUC & WAU & LIFT@30 & QINI & AUUC & WAU & LIFT@30 & Time(s) & Epochs & Params \\ 
     \hline
        
    \multirow{13}{*}{Criteo}  
        & 2012 & S-Learner & \underline{0.0988} & \underline{0.0387} & 0.0080 & 0.0306 & 0.0902 & \underline{0.0354} & 0.0078 & 0.0298 & 100 & 11 & 7705 \\ 
        & 2016 & T-Learner & 0.0888 & 0.0348 & 0.0078 & 0.0284 & 0.0790 & 0.0310 & 0.0077 & 0.0277 & 49 & 1 & 4314 \\ 
        & 2016 & BNN & 0.0979 & 0.0384 & 0.0078 & 0.0309 & 0.0888 & 0.0349 & 0.0077 & 0.0286 & 68 & 12 & 28697 \\ 
        & 2017 & TARNet & 0.0952 & 0.0374 & 0.0080 & 0.0308 & 0.0847 & 0.0333 & 0.0079 & 0.0288 & 215 & 15 & 3898 \\ 
        & 2017 & CFRNet & \underline{0.0994} & \underline{0.0390} & 0.0078 & \textbf{0.0317} & \underline{0.0901} & \underline{0.0354} & 0.0079 & \textbf{0.0302} & 139 & 15 & 14426 \\ 
        & 2017 & CEVAE & 0.0950 & 0.0373 & \textbf{0.0083} & 0.0306 & 0.0867 & 0.0341 & \textbf{0.0082} & 0.0290 & 389 & 10 & 155606 \\ 
        & 2018 & GANITE & 0.0850 & 0.0334 & 0.0072 & 0.0291 & 0.0818 & 0.0322 & 0.0072 & 0.0283 & 967 & 2 & 139933 \\ 
        & 2019 & DragonNet & 0.0947 & 0.0371 & \textbf{0.0083} & 0.0303 & 0.0851 & 0.0335 & \underline{0.0080} & 0.0289 & 281 & 1 & 72092 \\ 
        & 2021 & FlexTENet & \textbf{0.1015} & \textbf{0.0398} & 0.0080 & \textbf{0.0317} & \textbf{0.0924} & \textbf{0.0363} & 0.0077 & \underline{0.0298} & 475 & 9 & 36219 \\ 
        & 2021 & SNet & 0.0973 & 0.0382 & \textbf{0.0083} & \underline{0.0311} & 0.0843 & 0.0331 & \underline{0.0081} & 0.0284 & 237 & 4 & 27803 \\ 
        & 2021 & EUEN & 0.0949 & 0.0373 & 0.0079 & 0.0308 & \underline{0.0898} & 0.0353 & 0.0077 & \underline{0.0300} & 122 & 12 & 30756 \\ 
        & 2022 & DESCN & 0.0831 & 0.0327 & 0.0074 & 0.0284 & 0.0803 & 0.0316 & 0.0074 & 0.0274 & 234 & 10 & 98973 \\ 
        & 2023 & EFIN & 0.0949 & 0.0372 & 0.0077 & 0.0308 & 0.0859 & 0.0337 & 0.0074 & 0.0287 & 168 & 2 & 3523 \\ \hline

        \multirow{13}{*}{Lazada} 
        & 2012 & S-Learner & 0.4581 & 0.0030 & 0.0404 & 0.0568 & 0.0207 & 0.0029 & 0.0038 & 0.0072 & 14 & 1 & 75432 \\ 
        & 2016 & T-Learner & 0.4648 & 0.0036 & 0.0386 & 0.0549 & 0.0217 & 0.0031 & 0.0038 & \underline{0.0084} & 7 & 2 & 4615 \\ 
        & 2016 & BNN & 0.4558 & 0.0057 & 0.0416 & 0.0608 & \underline{0.0230} & \underline{0.0033} & 0.0039 & 0.0073 & 11 & 4 & 12391 \\ 
        & 2017 & TARNet & 0.4635 & 0.0053 & 0.0416 & 0.0612 & 0.0214 & 0.0030 & \textbf{0.0040} & 0.0072 & 19 & 4 & 6312 \\ 
        & 2017 & CFRNet & 0.4609 & 0.0078 & \underline{0.0421} & 0.0631 & \underline{0.0231} & \underline{0.0033} & 0.0038 & \textbf{0.0087} & 8 & 6 & 64680 \\ 
        & 2017 & CEVAE & 0.4734 & 0.0053 & \underline{0.0420} & 0.0564 & 0.0158 & 0.0022 & 0.0037 & 0.0074 & 18 & 7 & 72738 \\ 
        & 2018 & GANITE & 0.4592 & 0.0065 & 0.0181 & 0.0567 & 0.0174 & 0.0024 & 0.0036 & 0.0078 & 17 & 15 & 17003 \\ 
        & 2019 & DragonNet & \underline{0.5051} & \textbf{0.0196} & \textbf{0.0475} & \textbf{0.0800} & 0.0176 & 0.0025 & 0.0036 & 0.0077 & 9 & 9 & 23338 \\ 
        & 2021 & FlexTENet & 0.4576 & 0.0029 & 0.0385 & 0.0540 & 0.0185 & 0.0026 & \textbf{0.0040} & \underline{0.0079} & 8 & 1 & 49993 \\ 
        & 2021 & SNet & \underline{0.4767} & \underline{0.0110} & 0.0317 & 0.0582 & \textbf{0.0238} & \textbf{0.0034} & \textbf{0.0040} & 0.0076 & 12 & 15 & 97833 \\ 
        & 2021 & EUEN & 0.4758 & \underline{0.0108} & 0.0412 & \underline{0.0687} & 0.0156 & 0.0022 & 0.0038 & 0.0073 & 20 & 5 & 50010 \\ 
        & 2022 & DESCN & 0.4741 & 0.0056 & 0.0385 & 0.0579 & 0.0223 & 0.0031 & 0.0039 & 0.0075 & 9 & 5 & 9131 \\ 
        & 2023 & EFIN & \textbf{0.5131} & 0.0084 & 0.0388 & \underline{0.0655} & 0.0141 & 0.0020 & 0.0036 & 0.0077 & 83 & 1 & 376914 \\ 
        
        \specialrule{0.1em}{3pt}{3pt}
    \end{tabular}}
    \label{tab:result_3}
\end{table*}

\begin{table*}[htbp]
\caption{Benchmark results on Criteo and Lazada dataset without instance deduplication and with feature normalization. We highlight the top 3 best results, where the best results are marked in bold.}
    \centering
    \resizebox{0.89\textwidth}{!}{
    \begin{tabular}{c|c|c|cccc|cccc|ccc}
    \specialrule{0.1em}{3pt}{3pt}
    
     \multirow{2}*{Dataset} &
     \multirow{2}*{Year} & 
     \multirow{2}*{Model} & 
     \multicolumn{4}{c|}{Valid Metrics} & 
     \multicolumn{4}{c|}{Test Metrics} & 
     \multicolumn{3}{c}{Train Infos} \\
     \cline{4-14} 
       & & &QINI & AUUC & WAU & LIFT@30 & QINI & AUUC & WAU & LIFT@30 & Time(s) & Epochs & Params \\ 
     \hline
        
        \multirow{13}{*}{Criteo}  
        & 2012 & S-Learner & \textbf{0.1020} & \textbf{0.0400} & 0.0083 & \textbf{0.0312} & 0.0868 & 0.0341 & 0.0081 & 0.0288 & 100 & 8 & 7705 \\ 
        & 2016 & T-Learner & 0.0973 & 0.0382 & 0.0081 & \underline{0.0309} & 0.0843 & 0.0331 & 0.0081 & 0.0285 & 113 & 8 & 15258 \\ 
        & 2016 & BNN & \underline{0.0997} & \underline{0.0391} & 0.0082 & \underline{0.0309} & 0.0828 & 0.0326 & 0.0080 & 0.0282 & 76 & 10 & 28697 \\ 
        & 2017 & TARNet & 0.0981 & 0.0385 & 0.0079 & 0.0304 & 0.0834 & 0.0327 & 0.0078 & 0.0283 & 110 & 10 & 14426 \\ 
        & 2017 & CFRNet & 0.0984 & 0.0386 & 0.0083 & \underline{0.0309} & \underline{0.0892} & \underline{0.0351} & \underline{0.0083} & \textbf{0.0297} & 85 & 7 & 3898 \\ 
        & 2017 & CEVAE & 0.0713 & 0.0281 & \textbf{0.0084} & 0.0261 & 0.0660 & 0.0260 & \textbf{0.0084} & 0.0248 & 120 & 1 & 42878 \\ 
        & 2018 & GANITE & 0.0829 & 0.0326 & 0.0072 & 0.0285 & 0.0807 & 0.0318 & 0.0071 & 0.0278 & 890 & 1 & 10045 \\ 
        & 2019 & DragonNet & 0.0992 & \underline{0.0390} & 0.0082 & 0.0307 & 0.0886 & 0.0348 & 0.0081 & 0.0288 & 58 & 9 & 4988 \\ 
        & 2021 & FlexTENet & \underline{0.0994} & \underline{0.0390} & 0.0082 & 0.0306 & \underline{0.0894} & \underline{0.0351} & 0.0079 & 0.0290 & 484 & 15 & 138971 \\ 
        & 2021 & SNet & 0.0977 & 0.0383 & 0.0080 & 0.0307 & 0.0851 & 0.0334 & 0.0080 & 0.0283 & 78 & 7 & 125403 \\ 
        & 2021 & EUEN & 0.0987 & 0.0388 & \textbf{0.0084} & 0.0305 & \textbf{0.0899} & \textbf{0.0353} & 0.0082 & \underline{0.0295} & 74 & 1 & 15292 \\ 
        & 2022 & DESCN & 0.0827 & 0.0325 & 0.0069 & 0.0287 & 0.0803 & 0.0316 & 0.0070 & 0.0279 & 132 & 7 & 25437 \\ 
        & 2023 & EFIN & 0.0961 & 0.0377 & \textbf{0.0084} & 0.0308 & 0.0857 & 0.0337 & \underline{0.0083} & \underline{0.0294} & 586 & 1 & 21216 \\\hline 

        \multirow{13}{*}{Lazada} 
        & 2012 & S-Learner & 0.4288 & 0.0017 & 0.0355 & 0.0520 & 0.0218 & 0.0031 & 0.0036 & \underline{0.0081} & 13 & 15 & 37927 \\ 
        & 2016 & T-Learner & 0.4455 & 0.0046 & 0.0335 & 0.0542 & 0.0237 & \underline{0.0034} & 0.0040 & 0.0074 & 14 & 5 & 75432 \\ 
        & 2016 & BNN & 0.4688 & 0.0050 & 0.0401 & 0.0576 & 0.0184 & 0.0026 & \underline{0.0041} & 0.0074 & 11 & 3 & 4615 \\ 
        & 2017 & TARNet & 0.4381 & 0.0016 & 0.0353 & 0.0510 & 0.0195 & 0.0028 & 0.0040 & 0.0069 & 14 & 3 & 64680 \\ 
        & 2017 & CFRNet & 0.4624 & 0.0029 & 0.0407 & 0.0546 & 0.0219 & 0.0031 & \underline{0.0041} & \textbf{0.0082} & 10 & 5 & 6312 \\ 
        & 2017 & CEVAE & 0.4509 & 0.0069 & \underline{0.0425} & 0.0616 & 0.0220 & 0.0031 & 0.0036 & 0.0069 & 33 & 4 & 201330 \\ 
        & 2018 & GANITE & 0.4873 & 0.0067 & 0.0193 & 0.0566 & 0.0167 & 0.0023 & 0.0037 & 0.0075 & 34 & 15 & 50475 \\ 
        & 2019 & DragonNet & \underline{0.5103} & \textbf{0.0164} & 0.0459 & \textbf{0.0794} & 0.0204 & 0.0029 & 0.0038 & 0.0074 & 11 & 12 & 23338 \\ 
        & 2021 & FlexTENet & 0.4374 & 0.0030 & 0.0323 & 0.0510 & \underline{0.0238} & \underline{0.0034} & 0.0038 & 0.0076 & 14 & 3 & 93177 \\ 
        & 2021 & SNet & 0.4605 & 0.0091 & 0.0391 & 0.0586 & \textbf{0.0272} & \textbf{0.0039} & \textbf{0.0042} & 0.0075 & 9 & 4 & 157353 \\ 
        & 2021 & EUEN & 0.4758 & 0.0079 & 0.0402 & 0.0596 & \underline{0.0245} & \underline{0.0035} & 0.0038 & \underline{0.0079} & 9 & 3 & 50010 \\ 
        & 2022 & DESCN & \underline{0.5134} & \underline{0.0131} & \underline{0.0444} & \underline{0.0747} & 0.0185 & 0.0026 & 0.0036 & \underline{0.0079} & 10 & 2 & 108203 \\ 
        & 2023 & EFIN & \textbf{0.5364} & \underline{0.0135} & \textbf{0.0475} & \underline{0.0699} & 0.0163 & 0.0023 & 0.0036 & 0.0070 & 27 & 2 & 95198 \\ 
        
        \specialrule{0.1em}{3pt}{3pt}
    \end{tabular}}
    \label{tab:result_4}
\end{table*}

\subsection{Sensitivity of Instance Deduplication}
We first conduct two corresponding experiments to evaluate the impact of the instance deduplication operation on DUM, considering the case where the feature normalization operation is performed or not performed.

\subsubsection{No Feature Normalization Operation is Performed}
\begin{figure*}[htbp]
\centering
\includegraphics[width=0.9\textwidth]{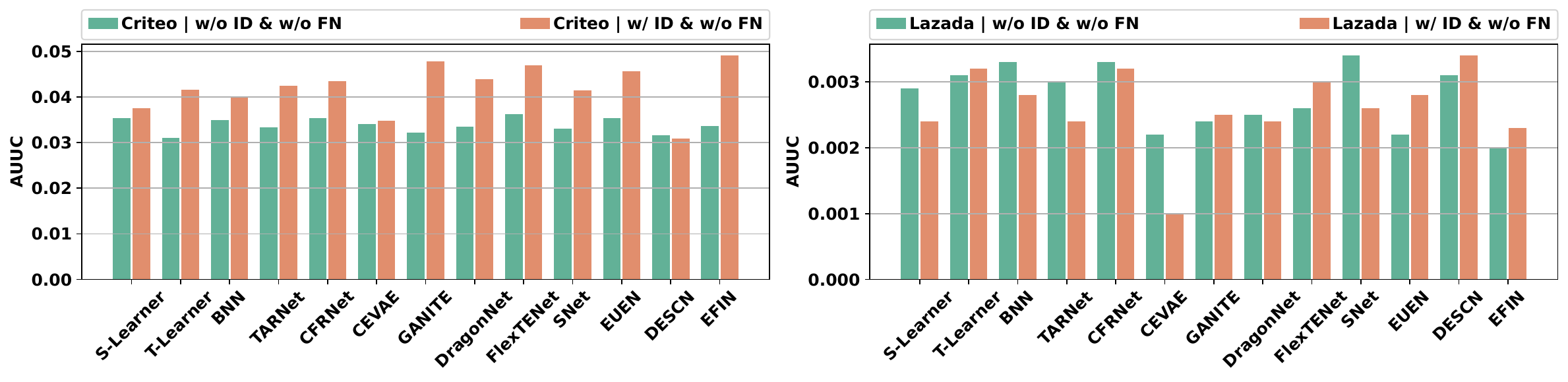}
\caption{The impact of whether to perform instance deduplication (i.e., ``w/ ID'' or ``w/o ID'') on two benchmark datasets when feature normalization is not performed (i.e., ``w/o FN'').}
\label{fig:2}
\end{figure*}

We report the corresponding results in Figure~\ref{fig:2}, where the left and right represent those obtained on Criteo and Lazada, respectively.
Based on the results of Figure~\ref{fig:2}, we can have the following observations:
1) Without performing feature normalization operations, by performing instance deduplication operations on Criteo, i.e., when there is no apparent inconsistency between the training distribution and the test distribution, the performance of the DUM model can be significantly improved in most cases, except for CEVAE and DESCN, and
2) However, by performing instance deduplication on Lazada, i.e., when there is an apparent inconsistency between the training distribution and the test distribution, most DUM models will suffer, although a small number of DUM models can benefit from it.

\begin{figure*}[htbp]
\centering
\includegraphics[width=0.9\textwidth]{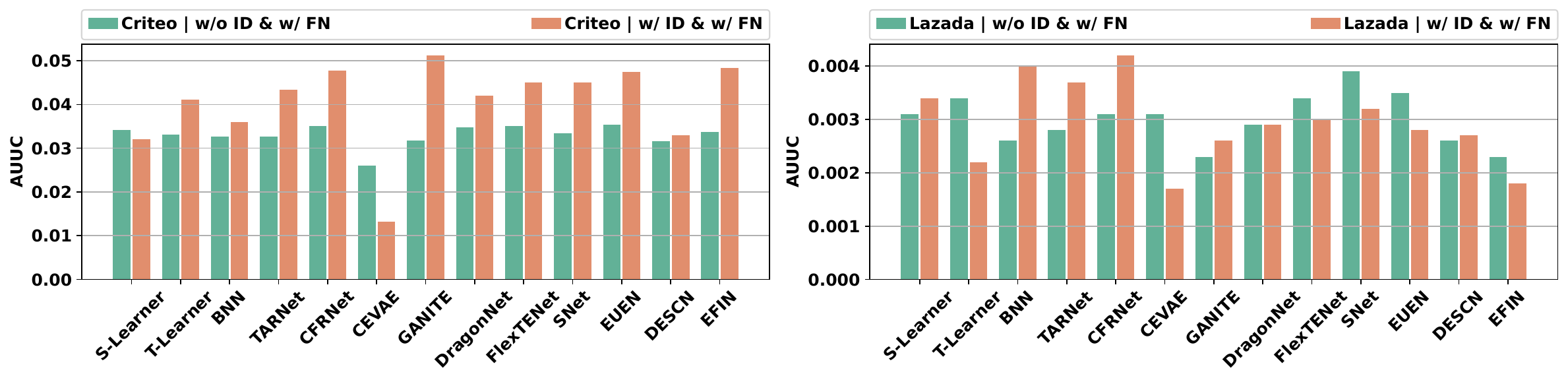}
\caption{The impact of whether to perform instance deduplication (i.e., ``w/ ID'' or ``w/o ID'') on two benchmark datasets when feature normalization is performed (i.e., ``w/ FN'').}
\label{fig:3}
\end{figure*}
\begin{figure*}[htbp]
\centering
\includegraphics[width=0.9\textwidth]{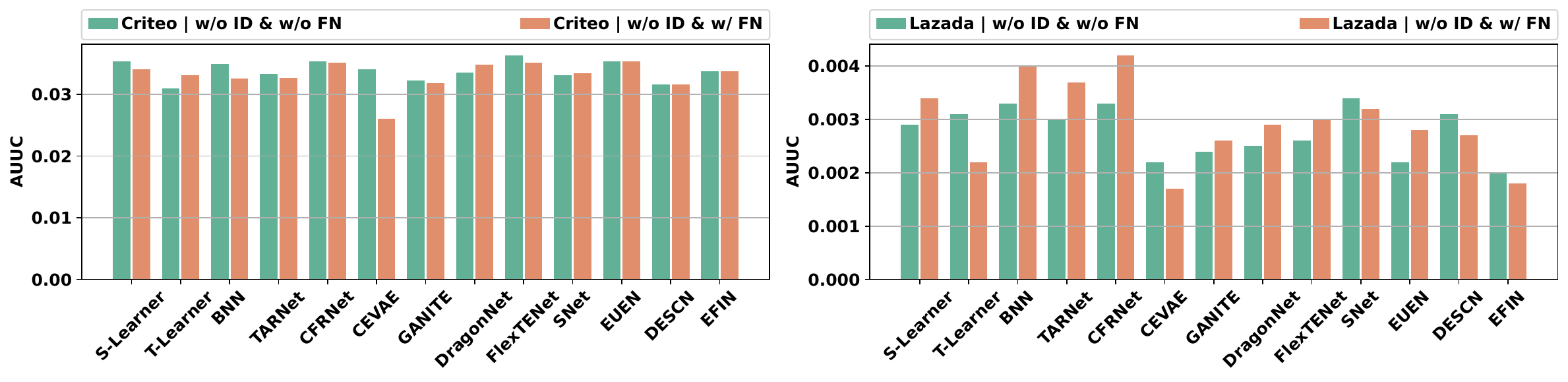}
\caption{The impact of whether to perform feature normalization (i.e., ``w/ FN'' or ``w/o FN'') on two benchmark datasets when instance deduplication is not performed (i.e., ``w/o ID'').}
\label{fig:4}
\end{figure*}
\begin{figure*}[htbp]
\centering
\includegraphics[width=0.9\textwidth]{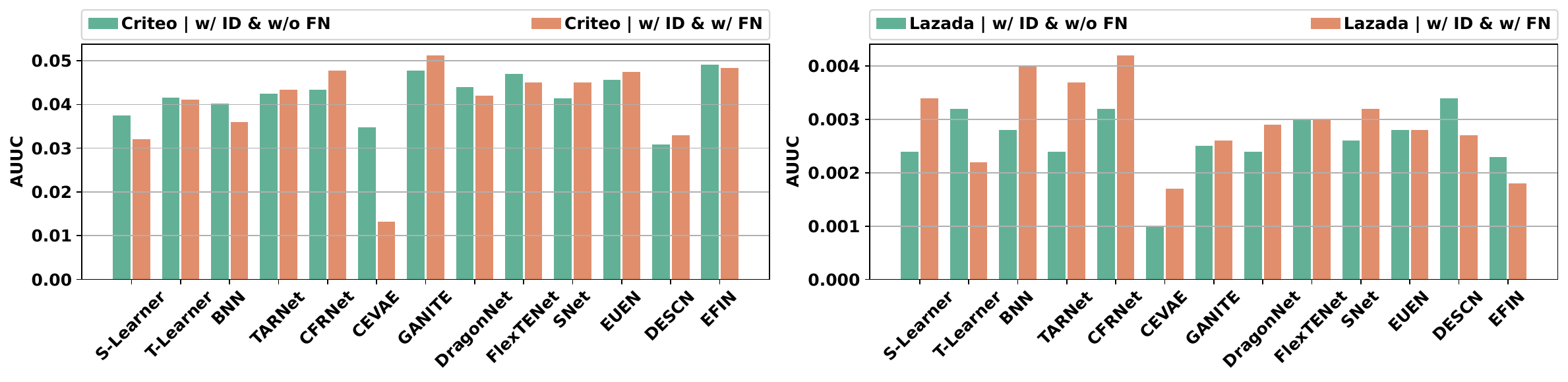}
\caption{The impact of whether to perform feature normalization (i.e., ``w/ FN'' or ``w/o FN'') on two benchmark datasets when instance deduplication is performed (i.e., ``w/ ID'').}
\label{fig:5}
\end{figure*}

\subsubsection{Feature Normalization Operation is Performed}
We report the corresponding results in Figure~\ref{fig:3}, where the left and right represent those obtained on Criteo and Lazada, respectively.
Based on the results of Figure~\ref{fig:3}, we can have the following observations:
1) Compared with the case without feature normalization (i.e., Figure~\ref{fig:2}), the trends of most DUM models on Criteo are similar, and
2) However, most DUM models on Ladaza will have an opposite trend compared to the case without feature normalization (i.e., Figure~\ref{fig:2}).

\subsubsection{Remark}
Based on the above results, we can find that when deploying a specific DUM model in practical applications, practitioners must consider its compatibility with duplicate instances to determine whether to perform instance deduplication.
In addition, it also shows that most DUM models are not robust to the presence of duplicate instances, which requires more research attention on data evaluation, denoising, and selection of DUM models.


\subsection{Sensitivity of Feature Normalization}
To evaluate the impact of the feature normalization operation on DUM, we conduct two corresponding experiments, respectively, considering the case where the instance deduplication operation is performed or not performed.

\subsubsection{No Instance Deduplication Operation is Performed}
We report the corresponding results in Figure~\ref{fig:4}, where the left and right represent those obtained on Criteo and Lazada, respectively.
Based on the results of Figure~\ref{fig:4}, we can have the following observations:
1) On Criteo, performing feature normalization operations has almost no obvious impact without performing instance deduplication operations, except for a slight decrease in the performance of some DUM models.
2) However, we can find that the impact of performing feature normalization operations on Ladaza is bipolar, with significant improvements to S-Learner, BNN, TARNet, and CFRNet but damage to others.


\subsubsection{Instance Deduplication Operation is Performed}
We report the corresponding results in Figure~\ref{fig:5}, where the left and right represent those obtained on Criteo and Lazada, respectively.
Based on the results of Figure~\ref{fig:5}, we can have the following observations:
1) Compared to the case where instance deduplication is not performed (i.e., Figure~\ref{fig:4}), the trends are similar on most DUMs on Criteo, except for more significant fluctuations on several models, such as CFRNet and GANITE.
2) The performance trend changes on Ladaza are similar to those observed on Criteo.

\subsubsection{Remark}
The above results suggest that practitioners also need to make more considerations in feature processing to determine the best operation when deploying a specific DUM model.
Furthermore, it also shows that most DUM models are not robust to different feature processing, requiring more research on feature selection and feature embedding learning of DUM.


\subsection{Sensitivity of Test Distribution}
By comparing the results in Tables~\ref{tab:result_1},~\ref{tab:result_2},~\ref{tab:result_3}, and~\ref{tab:result_4}, we can find that no matter which data preprocessing combination is used, most DUM models often have different performance on Criteo and Lazada.
Taking the results of a recent DUM model, EFIN, as an example, it has excellent performance in most cases on the Criteo dataset.
However, its performance on the Lazada dataset is particularly weak.
This primarily reflects that most existing DUM models are not robust enough when facing different test distributions, where Criteo has a selection-biased test environment, and Lazada has an unbiased test environment.
This requires practitioners to reasonably select an appropriate DUM model for deployment based on the nature of different business scenarios.
This can also motivate more attention to robust or out-of-distribution learning of DUM models and the automatic selection of the best DUM model in offline scenarios.



%% file: section/limitation.tex
\section{Limitations and Future Work}\label{sec:limitation}
In this section, we will address the limitations of our study and explore potential directions for future research.

\textbf{Dataset.}
In our open benchmark, we evaluated the performance of various uplift models using two real-world industrial datasets, Criteo~\cite{diemert2021large} and Lazada~\cite{zhong2022descn}. 
To the best of our knowledge, EC-LIFT~\cite{ke2021addressing} is another large-scale industrial dataset for different brands in a large-scale advertising scene, which is open-sourced by Alimama and contains billions of instances. 
We plan to extend more datasets to make it a more comprehensive open benchmark for uplift modeling.

\textbf{Instance and Feature Processing.}
In the previous sections, we discussed the impact of data normalization and instance deduplication on the accuracy of uplift prediction. However, other factors can also affect the performance of uplift modeling. In our subsequent research, we plan to offer a more comprehensive range of instance and feature processing methods to evaluate the impact of different data processing techniques on uplift prediction accuracy.

\textbf{Models.}
At present, our open benchmark only compares the performance of binary intervention uplift models. In our subsequent work, we plan to expand the benchmark to support multi-value intervention, continuous value intervention, and time-variant intervention uplift modeling. We also plan to add more representative models to further enhance its comprehensiveness and value.

%% file: section/conclusion.tex
\section{Conclusions}\label{sec:conclusion}
This paper presents the first open benchmark, DUMOM, for deep uplift modeling (DUM) tasks.
Our goal is to address the problem of non-reproducible and inconsistent results in this area of research.
We design a standardized evaluation protocol to reevaluate 13 existing models by employing four preprocessing settings on two widely used real-world datasets. 
We rigorously compare existing models and provide a fair benchmark result.
The results show that achieving substantial performance improvements in DUM studies is challenging and requires special consideration of generalization issues.
In addition, we provide some valuable experience when deploying specific deep boosting models, i.e., practitioners must consider more about the impact of different data preprocessing on specific DUM models.
We believe our benchmark helps drive further developments in this research area and for evaluating new studies' fair performance.